\documentclass[a4paper,conference]{IEEEtran}
\usepackage[left=1.57cm,right=1.57cm,top=0.95cm,bottom=2.7cm]{geometry}

\pdfpageattr{/Group << /S /Transparency /I true /CS /DeviceRGB >>}

\usepackage[utf8]{inputenc}
\usepackage[T1]{fontenc}
\usepackage{textcomp}
\usepackage{microtype}
\usepackage{calc}
\usepackage[normalem]{ulem}
\usepackage{verbatim}

\usepackage{lipsum}

\usepackage[range-phrase=--,per-mode=symbol-or-fraction,range-units=single,list-units=single,detect-all,forbid-literal-units=true]{siunitx}
\AtBeginDocument{
\DeclareSIUnit\bit{bit}
\DeclareSIUnit\byte{Byte}
\DeclareSIUnit\mbps{\mega\bit\per\second}
\DeclareSIUnit\kmh{\kilo\meter\per\hour}
\DeclareSIUnit\mw{\milli\watt}
\DeclareSIUnit\decibelm{dBm}
\DeclareSIUnit\decibeli{dBi}
\DeclareSIUnit\vehicle{veh}
}

\usepackage{silence}
\usepackage{csquotes}

\usepackage{amsmath}

\usepackage{amssymb}
\usepackage{amsfonts}
\usepackage{amsthm}
\usepackage{algpseudocode}

\usepackage{booktabs}
\usepackage{url}

\usepackage[inline]{enumitem}

\usepackage[caption=false,font=footnotesize]{subfig}
\usepackage[pdftex]{graphicx}
\DeclareGraphicsExtensions{.pdf,.png,.jpg}
\usepackage{tikz}
\usetikzlibrary{arrows}
\usetikzlibrary{calc}
\usetikzlibrary{chains}
\usetikzlibrary{scopes}

\usepackage[american]{babel}
\usepackage{hyphenat}

\usepackage{algorithm}
\usepackage[capitalize,noabbrev,nameinlink]{cleveref}

\RequirePackage{xstring}
\RequirePackage{xparse}
\RequirePackage[]{acro}
\makeatletter
\@ifpackagelater{acro}{2019/02/17}{
	\NewDocumentCommand\acrodef{mO{#1}mG{}}{\DeclareAcronym{#1}{short={#2}, long={#3}, foreign-plural={}, #4}}
}{
	\NewDocumentCommand\acrodef{mO{#1}mG{}}{\DeclareAcronym{#1}{short={#2}, long={#3}, #4}}
}
\makeatother

\acrodef{AI}{Artificial Intelligence}
\acrodef{BER}{Bit Error Rate}
\acrodef{CSI}{Channel State Information}
\acrodef{IoT}{Internet of Things}
\acrodef{LBT}{Listen-Before-Talk}
\acrodef{MCS}{Modulation and Coding Scheme}
\acrodef{NG-TCMS}{Next-Generation Train Control and Monitoring System}
\acrodef{NR}{New Radio}
\acrodef{QoS}{Quality of Service}
\acrodef{SCI}{Sidelink Control Information}
\acrodef{SCS}{Subcarrier Spacing}
\acrodef{SL}{Sidelink}
\acrodef{SNR}{Signal-to-Noise-Ratio}
\acrodef{SPS}{Semi-Persistent Scheduling}
\acrodef{RSRP}{Reference Signal Received Power}
\acrodef{TB}{Transport Block}
\acrodef{TSN}{Time-Sensitive Networking}
\acrodef{UE}{User Equipment}
\acrodef{V2I}{Vehicle-to-Infrastructure}
\acrodef{V2P}{Vehicle-to-Pedestrian}
\acrodef{V2V}{Vehicle-to-Vehicle}
\acrodef{V2X}{Vehicle-to-Everything}
\acrodef{WLCN}{WireLess Consist Network}
\acrodef{WLTB}{WireLess Train Backbone}
\acrodef{PRR}{Packet Reception Ratio}
\acrodef{PRB}{Physical Resource Block}
\acrodef{SINR}{Signal-to-Interference-plus-Noise-Ratio}
\acrodef{gNB}{g-NodeB}
\acrodef{LoS}{Line-of-Sight}

\acrodef{B.A.T.M.A.N.}{Better Approach To Mobile Ad-hoc Networking}
\acrodef{TTL}{Time to Live}
\acrodef{RB}{Resource Block}
\acrodef{OGM}{Originator Message}
\acrodef{MPR}{Multipoint Relaying}
\acrodef{PDR}{Packet delivery ratio}
\acrodef{D2D}{Device-to-Device}
\acrodef{OLSR}{Optimized Link State Routing Protocol}
\acrodef{AODV}{Ad hoc On-Demand Distance Vector Protocol}

\usepackage{todonotes}
\def\todoCtd#1{%
	TODO: #1%
	\ifx&#1&...\fi%
	\endgroup
	\relax
}

\NewDocumentCommand\IEEE{ s m >{\SplitArgument{4}{/}}d[] }{%
	\IfBooleanTF{#1}{}{IEEE\,}
	\nolinebreak[2]
	#2%
	\IfNoValueTF{#3}{%
	}{%
		\sommerIEEELettersSlashed#3%
	}%
}
\newcommand{\sommerIEEELettersSlashed}[5]{%
	\IfNoValueTF{#2}{%
	}{%
		\nolinebreak[3]
	}%
	#1%
	\IfNoValueTF{#2}{}{/#2}%
	\IfNoValueTF{#3}{}{/#3}%
	\IfNoValueTF{#4}{}{/#4}%
	\IfNoValueTF{#5}{}{/#5}%
}

\usepackage{pifont}

\begin{document}

\title{Reasoning Systems as Structured Processes: Foundations, Failures, and Formal Criteria}

\author{%
\IEEEauthorblockN{%
    Saleh Nikooroo and Thomas Engel
}%

\small{
    \texttt{%
	saleh.nikooroo@uni.lu, thomas.engel@uni.lu
    }
}
\\
}

\maketitle


\begin{abstract}
This paper outlines a general formal framework for reasoning systems, intended to support future analysis of inference architectures across domains. We model reasoning systems as structured tuples comprising phenomena, explanation space, inference and generation maps, and a principle base. The formulation accommodates logical, algorithmic, and learning-based reasoning processes within a unified structural schema, while remaining agnostic to any specific reasoning algorithm or logic system.
We survey basic internal criteria—including coherence, soundness, and completeness—and catalog typical failure modes such as contradiction, incompleteness, and non-convergence. The framework also admits dynamic behaviors like iterative refinement and principle evolution.
The goal of this work is to establish a foundational structure for representing and comparing reasoning systems, particularly in contexts where internal failure, adaptation, or fragmentation may arise. No specific solution architecture is proposed; instead, we aim to support future theoretical and practical investigations into reasoning under structural constraint.
\end{abstract}

\begin{IEEEkeywords}
Reasoning systems, formal structure, inference dynamics, failure modes, coherence, soundness, adaptive reasoning.
\end{IEEEkeywords}

\acresetall
\IEEEpeerreviewmaketitle

%


\section{Introduction}

Reasoning systems are typically formalized within well-established frameworks such as symbolic logic, optimization theory, or machine learning architectures. These paradigms often operate under the assumptions of full internal consistency, total deductive closure, or the global applicability of inference rules. While these assumptions enable tractable and mathematically elegant systems, they can obscure or fail to accommodate the real-world characteristics of reasoning: partial information, structural fragmentation, or evolving principle sets.

This paper departs from traditional models by proposing a general framework for reasoning systems as structured, modular entities. Crucially, our formulation does not presuppose any particular reasoning algorithm or deductive paradigm; it aims instead to describe the structural conditions under which reasoning may emerge or fail.

Rather than focusing solely on logical deduction or optimization performance, we examine reasoning systems as dynamic processes that interpret phenomena, generate explanations, and validate them internally against a set of governing principles.

Our motivation stems from the observation that many reasoning failures---from logical paradoxes to deadlocks in constrained systems---are not anomalies but structural symptoms of rigid or insufficient internal frameworks. To better understand these limits and potentials, we seek a formalism that:
\begin{itemize}
    \item Captures the internal composition of a reasoning system, including its inputs, outputs, generative maps, and principle base;
    \item Enables a classification of failure modes---such as contradiction, incompleteness, or non-convergence---within a unified framework;
    \item Provides structural criteria for internal coherence, inference validity, and explanation sufficiency;
    \item Views reasoning as an iterative and adaptive structural process, rather than a static rule-based engine.
\end{itemize}

The remainder of this paper proceeds as follows. In Section~\ref{sec:formal_model}, we define a minimal formal structure for reasoning systems. Section~\ref{sec:criteria} introduces key notions of coherence, soundness, and completeness. Section~\ref{sec:failures} categorizes failure modes and their structural implications. Section~\ref{sec:dynamics} examines the internal dynamics of reasoning, including fixed-point behavior and principle drift. Section~\ref{sec:examples} instantiates the formalism in several reasoning domains. We conclude with a brief discussion of implications and future directions.

%
\section{Formal Structure of a Reasoning System}
\label{sec:formal_model}

We define a reasoning system as a structured quintuple:
\[
\mathcal{R} = (P, E, f, g, \Pi)
\]
where each component plays a distinct functional role:

\begin{itemize}
    \item $P$ is the set of \textbf{phenomena}, inputs, or observed problems that the system is intended to interpret or solve;
    \item $E$ is the \textbf{explanation space}, consisting of candidate solutions, hypotheses, or structured outputs;
    \item $f : P \rightarrow E$ is the \textbf{inference map}, producing explanations from phenomena;
    \item $g : E \rightarrow P$ is the \textbf{generation map}, reconstructing or predicting phenomena from explanations;
    \item $\Pi$ is the \textbf{principle system}, a set of structural, logical, or epistemic constraints that govern both $f$ and $g$.
\end{itemize}

This formulation is intentionally agnostic to implementation. For instance:
\begin{itemize}
    \item In symbolic logic, $P$ may be a set of premises, $E$ a set of theorems, $f$ a deduction operator, and $\Pi$ a proof system;
    \item In constrained optimization, $P$ is the space of objective functions and constraints, $E$ the solution space, $f$ a solver, and $\Pi$ the feasibility conditions;
    \item In a neural inference setting, $f$ may be a learned function approximator, while $\Pi$ encodes architectural or regularization constraints.
\end{itemize}

We emphasize that $f$ and $g$ are not necessarily inverses, nor are they guaranteed to be bijective or even total. In fact, the limits of their definability and mutual consistency will be a central concern throughout this work.

The role of $\Pi$ is to constrain the admissibility and structural behavior of both $f$ and $g$. It may encode axioms, domain-specific rules, inductive biases, or dynamic constraints that evolve over time. Importantly, a reasoning system may be internally well-defined (i.e., $f$, $g$ exist and operate) while still violating $\Pi$, or failing to apply $\Pi$ consistently.

In what follows, we use this model to analyze the internal structure, coherence, and failure modes of reasoning systems in a general and implementation-independent way.

%

\section{Coherence, Soundness, and Validity}
\label{sec:criteria}

Having established the formal structure of a reasoning system $\mathcal{R} = (P, E, f, g, \Pi)$, we now define the core internal criteria by which such a system may be evaluated. These include:

\begin{itemize}
    \item \textbf{Coherence}: Whether the explanations produced by $f$ and their reconstructions via $g$ are consistent with the original inputs.
    \item \textbf{Soundness}: Whether all generated explanations respect the governing principles $\Pi$.
    \item \textbf{Completeness}: Whether the system is capable of producing valid explanations for all admissible phenomena in $P$.
\end{itemize}

\subsection{Coherence}
We define a reasoning system as \emph{coherent} if, for all $p \in P$ such that $f(p)$ is defined,
\[
g(f(p)) \approx p
\]
under a chosen notion of approximation (e.g., exact equality, semantic closeness, or acceptable reconstruction error). Coherence captures the ability of a reasoning system to reconstruct or validate its own interpretive steps.

Perfect coherence implies $g \circ f = \text{id}_P$ on the domain where $f$ is defined. In practice, this is often relaxed to a tolerable discrepancy, especially in probabilistic or learned systems.

\subsection{Soundness}
A reasoning system is \emph{sound} if all explanations produced by $f$ are consistent with the principles in $\Pi$. That is,
\[
\forall p \in P,\quad f(p) \models \Pi
\]
where $\models$ denotes logical or structural satisfaction. Soundness guarantees that inference respects internal rules, whether logical axioms, physical constraints, or algorithmic feasibility.

Soundness may also be localized: some systems are sound only with respect to subsets of $\Pi$, or under specific operational contexts.

\subsection{Completeness}
We say the system is \emph{complete} if, for every admissible $p \in P$, there exists a valid explanation $e \in E$ such that:
\[
f(p) = e \quad \text{and} \quad e \models \Pi
\]
This ensures that the reasoning system can handle its full intended domain without structural blind spots.

Note that completeness is not merely the existence of $f(p)$ but also its adherence to $\Pi$. A system may be able to compute $f(p)$ for all $p$, yet remain incomplete if those explanations violate $\Pi$.

\subsection{Fixed-Point Interpretations}
In certain systems, a fixed-point relationship may serve as an idealization:
\[
f(g(e)) = e \quad \text{and/or} \quad g(f(p)) = p
\]
Such relationships reflect ideal coherence under perfect invertibility. In general, systems approximate rather than achieve these fixed points.

\subsection{Joint Evaluation}
The interplay between coherence, soundness, and completeness reveals much about the system's structural health:

\begin{itemize}
    \item A system can be sound but incoherent (e.g., it reasons correctly but fails to reconstruct inputs).
    \item It can be coherent but incomplete (e.g., reconstructs reliably but fails on many inputs).
    \item Achieving all three properties simultaneously is rare and often requires tightly constrained $\Pi$ or system simplification.
\end{itemize}

These criteria provide a framework for analyzing where reasoning systems succeed or break down, setting the stage for the failure typologies to follow.

%

\section{Typology of Failure in Reasoning Systems}
\label{sec:failures}

Despite a system's formal structure and internal constraints, reasoning processes often encounter breakdowns. These failures are not necessarily flaws in implementation but rather indicators of structural insufficiency, misalignment, or rigidity within the system. We now outline a typology of such failures, categorized by their causes and manifestations.

\subsection{Contradiction}
A contradiction occurs when the output of the reasoning process violates the governing principles $\Pi$. That is, for some $p \in P$:
\[
f(p) \not\models \Pi
\]
This can arise from incompatible axioms, overloaded inference rules, or misaligned generative behavior in $g$. Contradictions indicate internal inconsistency and may lead to epistemic collapse or rejection of the explanation.

\subsection{Incompleteness}
A system is said to exhibit incompleteness when it fails to provide explanations for certain phenomena within its intended scope. Formally, there exists $p \in P$ such that $f(p)$ is undefined or inadmissible:
\[
\exists p \in P \quad \text{such that} \quad f(p) \notin E \quad \text{or} \quad f(p) \models \neg\Pi
\]
Incompleteness can stem from under-specified principles, rigid constraint boundaries, or unanticipated problem instances.

\subsection{Non-Convergence}
Some reasoning systems involve iterative or recursive processes. A failure mode emerges when these iterations do not converge to a stable explanation. For example, repeated application of $f \circ g$ or $g \circ f$ might yield divergent or oscillating results:
\[
\lim_{n \to \infty} (f \circ g)^n(e) \quad \text{does not exist}
\]
Non-convergence is common in optimization-based, heuristic, or neural reasoning frameworks.

\subsection{Overfitting and Underfitting}
When $f$ over-specializes to training or observed instances, the system may become brittle and fail to generalize—this is overfitting. Conversely, if $f$ is too coarse or regularized, it may produce vague or non-informative explanations—this is underfitting.

These behaviors often arise from misaligned $\Pi$, overly flexible or overly constrained function classes, or improper selection of inductive biases.

\subsection{Structural Deadlock}
A reasoning system may become structurally inert when its internal logic is self-consistent but incapable of progressing in the face of novel or ambiguous inputs. This deadlock occurs when:
\begin{itemize}
    \item $f$ is defined but constant or non-responsive across large regions of $P$;
    \item $\Pi$ prohibits admissible alternatives due to over-constraint;
    \item or $f$ outputs trivial explanations (e.g., $e = \texttt{null}$ or tautologies).
\end{itemize}

Such deadlocks are particularly insidious, as the system appears functional but fails to engage meaningfully with its problem space.

\subsection{Failure Summary}
These failure types are not mutually exclusive. A reasoning system may simultaneously suffer from contradiction in some regions, incompleteness in others, and deadlock elsewhere. The typology above provides a diagnostic vocabulary for characterizing and comparing reasoning systems in practice.

In the following section, we explore how such systems evolve internally—sometimes recovering from failure, sometimes reinforcing it.

%

\section{Internal Dynamics and Reasoning Evolution}
\label{sec:dynamics}

Reasoning systems are not always static entities. In many applications, they operate over time—through iterative refinement, self-correction, or evolving structural constraints. This section explores the internal dynamics of such systems, focusing on how they respond to error, adapt to novelty, or restructure themselves without external intervention.

\subsection{Sequential Inference and Iterative Structure}
In systems where $f$ or $g$ are defined recursively or in stages, reasoning proceeds through a sequence of internal steps:
\[
e_0 = f(p),\quad e_1 = f(g(e_0)),\quad \ldots,\quad e_n = f(g(e_{n-1}))
\]
This chain may converge, cycle, or diverge depending on the nature of $f$, $g$, and $\Pi$. The goal of such iterations may include refining an explanation, validating internal coherence, or escaping suboptimal initial mappings.

\subsection{Error-Driven Adjustment}
Some systems adjust their internal mappings in response to discrepancies between predicted and observed phenomena:
\[
\delta_p = p - g(f(p))
\]
Such error signals can trigger refinement of $f$, tuning of $g$, or revision of $\Pi$. This dynamic is especially prominent in adaptive learning systems, where gradient-based or rule-based updates aim to reduce reconstruction error or improve inference fidelity.

\subsection{Principle Drift}
Over time, the principle set $\Pi$ may itself evolve:
\[
\Pi_0 \rightarrow \Pi_1 \rightarrow \cdots \rightarrow \Pi_t
\]
This evolution may be triggered by contradictions, poor performance, or the emergence of new problem domains. Principle drift alters the admissibility conditions and effectively changes the structure of the reasoning system. It reflects a meta-level response: rather than repairing $f$ or $g$, the system alters what it considers valid.

\subsection{Self-Regularization}
Some reasoning systems include built-in mechanisms to avoid or correct undesirable behavior. These may include:
\begin{itemize}
    \item Penalizing incoherent mappings (e.g., regularization terms in optimization);
    \item Constraining search space to prevent overfitting;
    \item Disabling unstable regions in $f$ or $g$ through gating or pruning mechanisms.
\end{itemize}
These structural safeguards help enforce stability and steer the system toward valid reasoning behavior without explicit supervision.

\subsection{Local vs. Global Adaptation}
Adaptation may occur locally (e.g., only for specific phenomena $p \in P$) or globally (altering $f$, $g$, or $\Pi$ across the entire system). Systems that support local updates may be more resilient but risk fragmentation; global adaptations offer coherence at the risk of rigidity.

\subsection{Failure Response Modes}
Not all systems respond to failure. Some ignore error entirely; others collapse or halt; others adapt. The presence or absence of failure-response dynamics—especially principle drift and error correction—can be used to classify systems into static vs. evolving reasoning architectures.

In the next section, we instantiate this framework with examples drawn from logic, optimization, and learning systems.

%

%

\section{Examples of Reasoning Systems}
\label{sec:examples}

We now illustrate the general framework by instantiating it in three distinct domains: deductive logic, constrained optimization, and structured neural inference. These examples demonstrate the flexibility of the $(P, E, f, g, \Pi)$ formulation and how different reasoning paradigms fit within its structure.

\subsection{Example 1: Deductive Logic System}
\begin{itemize}
    \item \textbf{Phenomena} ($P$): Sets of premises or assumptions.
    \item \textbf{Explanation space} ($E$): Theorems or derived propositions.
    \item \textbf{Inference map} ($f$): A derivation function applying inference rules (e.g., modus ponens).
    \item \textbf{Generation map} ($g$): Reconstructs minimal premises or antecedents from a given theorem (where applicable).
    \item \textbf{Principles} ($\Pi$): Axioms, inference rules, and proof constraints (e.g., propositional logic axioms).
\end{itemize}

\textit{Coherence} in this setting corresponds to whether derived conclusions can be traced back to accepted premises. \textit{Soundness} ensures that all derivations respect logical axioms. \textit{Completeness} refers to whether all logically entailed theorems can be reached.

\subsection{Example 2: Constrained Optimization Solver}
\begin{itemize}
    \item \textbf{Phenomena} ($P$): Problem specifications—objective functions and constraint sets.
    \item \textbf{Explanation space} ($E$): Candidate solutions or configurations.
    \item \textbf{Inference map} ($f$): Optimization routine mapping problems to solutions.
    \item \textbf{Generation map} ($g$): Reconstructs problem structure from candidate solutions (e.g., via duality).
    \item \textbf{Principles} ($\Pi$): Feasibility conditions, KKT constraints, or convexity assumptions.
\end{itemize}

Here, \textit{coherence} implies that optimal solutions explain the posed problem faithfully. \textit{Soundness} requires that outputs satisfy all constraints. \textit{Completeness} reflects the solver’s ability to find feasible solutions across the problem domain.

\subsection{Example 3: Structured Neural Inference}
\begin{itemize}
    \item \textbf{Phenomena} ($P$): Input data points or observations (e.g., images, text).
    \item \textbf{Explanation space} ($E$): Feature embeddings, latent codes, or predicted labels.
    \item \textbf{Inference map} ($f$): Neural network performing encoding or classification.
    \item \textbf{Generation map} ($g$): Decoder or generative model reconstructing input.
    \item \textbf{Principles} ($\Pi$): Inductive biases encoded by architecture, loss functions, or regularization terms.
\end{itemize}

\textit{Coherence} relates to how well reconstructions match inputs. \textit{Soundness} reflects whether predictions respect the model’s inductive assumptions. \textit{Completeness} captures whether the model generalizes across the full data distribution.

\subsection{Cross-Example Summary}
Each of the above systems instantiates the same general structure but emphasizes different components of reasoning:

\begin{itemize}
    \item Logic emphasizes $\Pi$ (axioms) and $f$ (deduction).
    \item Optimization emphasizes $f$ and $g$ under feasibility constraints.
    \item Neural inference emphasizes learned $f$, approximate $g$, and implicit $\Pi$ via architecture.
\end{itemize}

This diversity underscores the versatility of the formal framework, which abstracts away from domain specifics to analyze structural and functional integrity at a higher level.

In the concluding section, we summarize our contributions and suggest directions for further research.

%

\section{Related Work}

\subsection*{Structured Reasoning Frameworks and Epistemic Models}

Recent work moves beyond classical logic by modeling reasoning as structured, modular, or coherence-driven rather than truth-functional. Simon \cite{simon2023biased} develops a coherence-based framework for biased reasoning, emphasizing internal consistency as the main organizing principle—resonant with our model’s allowance for structural failure and belief reconfiguration.

Casini \cite{casini2022conditional} provides a flexible logic for conditional reasoning beyond truth-functional semantics, supporting the kind of partial inference and non-monotonicity that our framework tolerates explicitly. Arieli \cite{arieli2023postulate} classifies argumentation schemes via postulate satisfaction, proposing inferential formalisms that account for contradiction without collapse, conceptually aligning with our treatment of epistemic tensions.

Gärdenfors \cite{gardenfors2023concepts} offers a geometric model of conceptual reasoning, and Kido \cite{kido2022generative} presents probabilistic symbolic frameworks for structured inference—both motivating our belief graph’s topological encoding of epistemic structure. Paulino-Passos \cite{paulino2022interactive} interprets explanation as a non-monotonic reasoning process, suggesting evolving inference paths; by contrast, we permit persistent contradiction to co-exist with structural coherence.

Bushuev \cite{bushuev2025triz} discusses multimodal reasoning via TRIZ principles, prioritizing adaptability and contradiction management—reflecting our interest in epistemic resilience rather than deductive purity.

\subsection*{Modular, Compositional, and Iterative Architectures}

The modular design of reasoning architectures has gained traction. Christianos et al. \cite{christianos2023pangu} propose Pangu-Agent, a generalist agent embedding reasoning modules for improved generalization. Creswell \cite{creswell2022faithful} builds modular chains of fine-tuned neural units to enforce reasoning faithfulness and reduce hallucination, paralleling our interest in structured yet flexible inference models.
SCREWS \cite{sridhar2023screws} introduces a reasoning framework that supports revision and modular recombination of inference steps. WanJun et al. \cite{wanjun2022reasonformer} disentangle representation and reasoning through a compositional transformer model, while Zhou et al. \cite{zhou2024selfdiscover} introduce SELF-DISCOVER to enable LLMs to autonomously compose reasoning chains—both reinforcing structural modularity over end-to-end black-box reasoning.
Fu \cite{fu2023morse} offers MORSE, a dynamic modular reasoning framework tailored to explanation generation. Hua et al. \cite{hua2022system1system2} combine neural and symbolic reasoning pathways in a two-system model, advocating for cognitive complementarity—a principle echoed in our separation of operational and constraint mappings.

\subsection*{Topological and Graph-Based Representations}

Graphical and topological reasoning models provide an interpretive scaffold. Zhang et al. \cite{zhang2024dot} propose Diagram of Thought (DoT), using Topos Theory to model LLM reasoning as a directed acyclic graph. Ho et al. \cite{ho2025dialogue} frame logical traversal as dialectical dialogue trees tolerant to inconsistency. Zhu et al. \cite{zhu2024structures} propose structural representation learning to generalize across reasoning types.

Our own belief graphs continue this lineage but shift emphasis from entailment to structural coherence—allowing persistent contradiction and cluster-based epistemics.

\subsection*{Meta-Reasoning, Verification, and Dynamic Strategy Control}

Sui \cite{sui2025meta} proposes Meta-Reasoner, a dynamic controller for strategy selection during inference time. Xiang \cite{xiang2025meta} extends this to meta chain-of-thought supervision, enhancing internal traceability.
Raza \cite{raza2025ssv} explores logical task verification using solvers, while ReCEval \cite{prasad2023receval} evaluates reasoning chains for correctness and informativeness. Both highlight meta-level oversight and structural inspection—similar in spirit to our separation of confidence versus credibility within belief networks.
Ling et al. \cite{ling2023deductive} offers a deductive framework supporting self-verification and chain decomposition, reinforcing reasoning transparency. Wei \cite{wei2025alignrag} proposes AlignRAG, a feedback-based framework for aligning multi-hop retrieval chains—conceptually related to our epistemic feedback model.

\subsection*{Inductive, Causal, and Failure-Oriented Models}

Qiu et al. \cite{qiu2023phenomenal} analyze inductive hypothesis refinement, revealing limitations in rule application and generalization—an issue our model reframes as structural incoherence. Wang et al. \cite{wang2024grokking} examine transformer generalization failures as indicators of latent reasoning limits.
Tang et al. \cite{tang2023towards} present CausalGPT, a multi-agent architecture for causality-aware reasoning. Saparov \cite{saparov2023greedy} offers a formal critique of LLM proof planning, demonstrating accurate micro-inference but poor global coherence. Jung et al. \cite{jung2022maieutic} improve reasoning robustness through recursive, logically consistent explanations, reinforcing our emphasis on iterative reasoning integrity.
Prystawski \cite{prystawski2023locality} proposes that reasoning arises from local correlations among observed variables—a structural insight compatible with our model’s cluster-driven reasoning across graph components.

\subsection*{General Syntheses and Reasoning Surveys}

Sun et al. \cite{sun2023survey} survey reasoning with foundation models, covering techniques across multi-modal, multi-agent, and structured reasoning. Plaat \cite{plaat2024survey} consolidates prompting and stepwise reasoning in LLMs, identifying current gaps in systematic reasoning.
Saied \cite{saied2024decision} surveys decision-making frameworks, emphasizing modular decision support—underscoring the need for structural substrates like belief graphs. Wang et al. \cite{wang2023boosting} and Radhakrishnan et al. \cite{radhakrishnan2023question} develop Chain-of-Knowledge prompting and decomposition-based reasoning, respectively, both contributing to the understanding of structured deduction.

\subsection*{Summary}

Across this landscape, a convergence emerges: toward frameworks that prioritize internal structure, revision, and epistemic transparency. Whether through modular reasoning units, topological encoding, or verification protocols, the field is moving away from opaque statistical inference toward explainable, compositional, and resilient systems. Our proposed belief graph framework fits within this shift, offering a contradiction-tolerant and structurally explicit substrate that separates reasoning integration from source reliability and procedural logic.

%

\section{Conclusion and Outlook}
\label{sec:conclusion}

This paper has proposed a general framework for reasoning systems, formalized as structured entities composed of phenomena, explanations, inference and generation mappings, and principle sets. We have shown that this formulation:

\begin{itemize}
    \item Captures diverse reasoning paradigms—including logic, optimization, and learning—under a unified structural model;
    \item Defines internal evaluation criteria such as coherence, soundness, and completeness;
    \item Supports a detailed typology of failure modes, including contradiction, incompleteness, non-convergence, and deadlock;
    \item Enables the study of internal dynamics, from iterative refinement and error correction to evolving principles and self-regularization.
\end{itemize}

By decoupling reasoning from any specific implementation or representational formalism, the model invites cross-domain analysis of structural reasoning properties. It offers a foundation for diagnosing reasoning pathologies, comparing systems, and guiding principled design.

\subsection*{Future Directions}
This work opens several avenues for future exploration:

\begin{itemize}
    \item \textbf{Modular Composition}: How can reasoning systems be composed from subsystems while preserving or enhancing structural integrity?
    \item \textbf{Resilience and Repair}: Can local adaptations (e.g., error-driven refinement) ensure global soundness and coherence?
    \item \textbf{Dynamic Principle Systems}: What governs the evolution of $\Pi$ in adaptive or self-revising systems?
    \item \textbf{Evaluation Metrics}: Beyond coherence and soundness, what metrics capture the robustness, generality, or strategic capacity of a reasoning system?
\end{itemize}

While we have not addressed interaction between distinct reasoning systems in this work, such considerations lie beyond the present scope and are left for future investigation.

\bigskip

\noindent In closing, the reasoning system framework offers a structural and dynamic view of inference architectures. It emphasizes not only what systems conclude, but how they operate, evolve, and fail. We believe this perspective will prove valuable across fields that rely on principled but adaptable reasoning processes.

%


\bibliography{main}

\end{document}